\documentclass[10pt,twocolumn,letterpaper]{article}
\usepackage{iccv}
\usepackage{times}
\usepackage{epsfig}
\usepackage{graphicx, subcaption}
\usepackage{svg}
\usepackage{amsmath}
\usepackage{amssymb}
\usepackage{comment}
\usepackage{tabularx,adjustbox,multicol,multirow,booktabs}

\usepackage[breaklinks=true,bookmarks=false]{hyperref}

\iccvfinalcopy 

\def\iccvPaperID{****} 

\ificcvfinal\fi

\begin{document}

\title{Deep Metric Learning-Based Out-of-Distribution Detection with Synthetic Outlier Exposure}

\author{Assefa Seyoum Wahd}

\maketitle

\begin{abstract}
In this paper, we present a novel approach that combines deep metric learning and synthetic data generation using diffusion models for out-of-distribution (OOD) detection. One popular approach for OOD detection is outlier exposure, where models are trained using a mixture of in-distribution (ID) samples and ``seen" OOD samples. For the OOD samples, the model is trained to minimize the KL divergence between the output probability and the uniform distribution while correctly classifying the in-distribution (ID) data. In this paper, we propose a label-mixup approach to generate synthetic OOD data using Denoising Diffusion Probabilistic Models (DDPMs). Additionally, we explore recent advancements in metric learning to train our models.

In the experiments, we found that metric learning-based loss functions perform better than the softmax. Furthermore, the baseline models (including softmax, and metric learning) show a significant improvement when trained with the generated OOD data. Our approach outperforms strong baselines in conventional OOD detection metrics.
\end{abstract}

\section{Introduction}

Out-of-distribution (OOD) detection plays a critical role in the development of robust machine learning models. While accurate classification of known classes is important, the ability to identify samples that deviate from the training distribution is equally crucial. This paper presents a novel approach that combines deep metric learning and synthetic data generation using diffusion models to improve OOD detection in classification models.

One popular approach for OOD detection is outlier exposure \cite{hendrycks2018deep}, which involves training models using seen samples from out-of-distribution data.The model is trained to output a low confidence on the training OOD data while correctly classifying the in-distribution (ID) data. Outlier exposure methods differ in how the OOD data is obtained (real-world data vs. generated data).

\cite{hendrycks2018deep} is the first successful work to train a classifier using labeled ID data and a large set of unlabeled OOD data. The method hypothesizes that training on a large and diverse OOD data can help deep neural networks (DNNs) generalize better to unseen OOD examples at test time. For example, they train a CIFAR-10 vs. others OOD detector by exposing the model with 80 Million Tiny Images dataset. Outlier exposure outperforms several state-of-the-art OOD detection methods on several benchmark datasets.

However, using a large unlabeled dataset as OOD training data introduces an unwanted problem.  Ideally, the training OOD data is considered to have no semantic similarity with the ID data. However, in practice, it is evident that OOD datasets obtained from the wild may contain mixed ID and OOD samples, thus introducing difficulty for outlier exposure methods that use large unlabeled OOD data for training. If the unlabeled dataset contains samples with overlapping semantics with the ID dataset, the network may just focus on minor statistical differences in the images and not the semantic meaning of the images. This may not be desirable as the model can easily overfit to the training ID and OOD datasets. To address this issue, \cite{yang2021semantically} trains an OOD detector by removing any overlapping classes from the OOD dataset by deep clustering \cite{caron2019deep}. \cite{katzsamuels2022training} models the mixed dataset as Huber contamination model \cite{huber1992robust}, meaning it is considered to be partially coming from ID and OOD distributions. The model is trained to predict the mixing ratio. Another challenge of utilzing real-world OOD data for training is that that data may not cover the full range of OOD examples that the model may encounter in the real world. Thus, training on OOD data may lead to overfitting on the specific OOD examples used for training. Furthermore, collecting real-world OOD data can be costly, for example, in case of rare events.

Other works have tried to solve the issue of overlapping training ID and OOD semantics and overfitting by generating synthetic OOD data that satisfy certain requirements \cite{lee2018training}, \cite{mirzaei2022fake}. The main idea is to generate samples that lie in the low density areas of the training data distribution; i.e., the generated data should be neither too close nor too far from the training data distribution. OOD data that is too close to the training distribution can limit the classifier’s closed-set classification accuracy, and data that is too far from the training distribution expands the classifier’s decision boundary, possibly classifying OOD datasets as ID. 

\begin{figure*}[!tb]
    \centering
    \includegraphics[width=\textwidth]{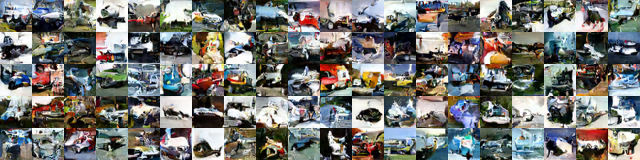}
    \caption{Synthetic OOD data generated using label mixup between CIFAR-10 ``airplane" and ``automobile" classes. The generated data have a significant diversity and meaningful mixup semantics. For example, a mixup between an airplane class and a automobile class results in an object with features from airplane and automobile.}
    \label{fig:diffusion_label_mixup}
\end{figure*}

In this paper, we propose a synthetic OOD data generation approach using denoising diffusion probabilistic Models (DDPMs) \cite{ho2020denoising}. Although this is not the first time diffusion models have been used for out-of-distribution detection, our method differs from previous works in several ways as will be explained next. \cite{mirzaei2022fake} generates synthetic OOD data by early stopping (i.e., before the model converges) a diffusion model during training. The authors argue that the Fréchet Inception Distance (FID) \cite{heusel2017gans}, which measures the quality of the generated image, has a direct correlation to the number of training steps and therefore early stopping ensures that the generated image doesn't fully converge to the training data distribution. The generated data is used to train a binary classifier and the nearest neighbor distance \cite{sun2022out} is used as the OOD score. \cite{graham2023denoising} proposed a reconstruction-based novelty detection by first adding a range of noise levels to a given input and then reconstructing it using a pre-trained DDPM. The reconstruction errors between the original and the reconstructed images are computed at several timesteps, and the average reconstruction error is used to classify an input as ID or OOD. \cite{liu2023unsupervised} used the image inpainting power of diffusion models as the OOD score function. Specifically, a test image is first corrupted by masking a large portion and a diffusion model is used to reconstruct the corrupted image. In-distribution samples have a small reconstruction error (because DDPMs have the ability to inpaint) while the OOD samples typically have a large reconstruction error. We propose a synthetic OOD generation approach by interpolating the one-hot encoding of the target classes in a conditional diffusion model. We refer to this approach as a label mixup. Suppose that we want to a cat vs. dog classifier. First, we train a class-conditional DDPM by giving the one-hot labels (i.e, $[1,0]$, and $[0,1]$) and a noise image to the DDPM. To generate a synthetic OOD data, we mix the labels by interpolating the one-hot encodings, i.e., $[1,1]$. Since, the generative models has only seen $[1, 0]$ or $[0, 1]$ during training, $[1,1]$ input generates data that lies between the two classes. See Figure \ref{fig:diffusion_label_mixup}. It is important to note the previous methods that use DDPM treat OOD detection as a binary classification hence do not train a multi-class classifier which is an important distinction to our work. 

It is a standard practice to train multi-class classification models with  the softmax loss function. In this work, we take inspiration from the success of contrastive learning methods in OOD detection to investigate alternative loss functions to train our OOD detection models. Contrastive learning \cite{guille2022cadet}, \cite{tack2020csi} trains a similarity function (e.g., cosine similarity \cite{guille2022cadet}, \cite{tack2020csi}) to maximize the similarity between different (``weak") augmentations of a given sample, and minimize the similarity with other samples (i.e., instance discrimination \cite{wu2018unsupervised}). Specifically, the goal in contrastive learning is to train an encoder neural network such that different random augmentations of the same image are close in the embedding space but far from the embeddings of another image. CSI \cite{tack2020csi} found that, in addition to pushing away different samples, pushing ``strong" augmentations of a sample (e.g., rotation) away from the original sample improves OOD detection as strong augmentation can shift the distribution of an input. To train a model using a contrastive loss function, negative sample mining that chooses for the most useful samples is crucial. In CSI, positive samples of an input are obtained with weak augmentations (e.g., cropping) while negative samples include strong augmentations of the same image as well as other images from the training dataset. CADet \cite{guille2022cadet} utilized the maximum mean discrepancy (MMD) two-sample test \cite{gretton2012kernel} as a score function for models trained with contrastive loss functions. Angle-based  metric learning methods (\cite{liu2017sphereface},\cite{wang2018cosface},\cite{zhang2019adacos},\cite{deng2020sub}), on the other hand, propose a similarity learning mechanism without negative sample mining. Metric (distance) learning techniques are commonly employed to increase inter-class variation and reduce intra-class variation in the feature space of deep neural networks, especially in few-shot settings \cite{sung2018learning},\cite{li2020bsnet},\cite{zhang2021rethinking},\cite{yang2020dpgn} and deep face recognition \cite{liu2017sphereface},\cite{wang2018cosface},\cite{zhang2019adacos},\cite{deng2020sub}. In this work, we regard state-of-the-art metric learning loss functions,  such as SphereFace \cite{liu2017sphereface}, CosFace \cite{wang2018cosface}, AdaCos \cite{zhang2019adacos}, and ArcFace \cite{deng2020sub} as OOD score functions.

To evaluate the effectiveness of our method, we compare it with well-known approaches in the field. Our results show that our approach outperform baseline methods in conventional OOD detection metrics (AUROC and AUPR). By combining deep metric learning and synthetic data generation, our proposed method offers a promising solution for improving OOD detection.

In summary the contributions of this paper are,
\begin{itemize}
    \item We introduce a synthetic out-of-distribution data generation using denoising diffusion models.
    \item We adapt popular loss functions in deep metric learning for out-of-distribution detection.
    \item We show that models trained with the proposed outlier exposure outperform the regular softmax and metric learning loss function.
    \item To the best of our knowledge this is the first time diffusion models have been used to generate synthetic OOD data by label mixup and used to train a multi-class classifier.
\end{itemize}

\section{Related Work}
\textit{Maximum-Softmax Probability (MSP) \cite{hendrycks2016baseline}.} This baseline method uses the highest output probability as the score function. The intuition is that a classifier should be more confident about in-distribution inputs than OOD inputs. The MSP score function is defined as follows:
    \begin{equation}
        s_\theta(x) = \max_{c \in C} p_\theta(y=c|x).
    \end{equation}
\textit{Energy-Based OOD Detection (EBO) \cite{liu2020energy}.} In EBO, an energy score is derived as the `logsumexp' of the output predictions scaled by a temperature $T$:
\begin{equation}
    s_\theta(x) = T\log\sum_i^C \exp \left(f_\theta(x; i) / T\right)
\end{equation}
where $f_\theta(x; i)$ is the logit value corresponding the $i-th$ class of the classifier $f_\theta$.





\textit{Mahalanobis Distance \cite{lee2018simple}.}
The Mahalanobis distance measures of how far a point is from the mean of a distribution. Firstly, class-conditional Gaussian distributions are formed from the features of the penultimate layer, with  $\mu_c = \frac{1}{N_c} \underset{i:y_i=c}{\sum^{N_c}} f_\theta(x_i)$, for $c=1, \dots, C$, a covariance matrix, $\Sigma = \frac{1}{N_c} \underset{c=1}{\sum^C}\underset{i:y_i=c}{\sum} (f_\theta(x_i) - \mu_c)(f_\theta(x_i) - \mu_c)^T$. Then the OOD score function, $s_\theta(x)$, is defined as the negative of the minimum distance from each conditional Gaussian distribution:
\begin{equation}
    s_\theta(x) = -\underset{c \in C}{\min}\; (f_\theta(x_i) - \mu_c)\Sigma^{-1}(f_\theta(x_i) - \mu_c)^T
\end{equation}

\textit{Outlier Exposure}. \cite{hendrycks2018deep} trains a classifier using labeled ID data and a large set of unlabeled OOD data. The models have better calibration and OOD detection ability. The problem of overlapping semantics between the training ID data and OOD data has been studied in \cite{yang2021semantically}. The method uses deep clustering \cite{caron2019deep} to filter out the semantically overlapping samples from the unlabeled OOD data. \cite{katzsamuels2022training} models the training OOD dataset as Huber contamination model \cite{huber1992robust}, meaning it is considered to be partially coming from ID and OOD distributions. The model is trained to predict the mixing ratio. \cite{fort2021exploring} studies the effectiveness of pre-trained transformer models for out-of-distribution detection. Their findings demonstrate that large scale pre-trained transformer models fine-tuned on the ID data have excellent discriminative ability, but not a well-separated boundary for OOD detection. To improve the OOD detection capability of such models the authors fine-tune the model on seen OOD samples. The key takeway is that setting aside a training OOD data is important to detect OOD samples at test time.

\textit{Synthetic Data Outlier Exposure}. Synthetic outlier exposure uses generated data as the seen OOD data. \cite{lee2018training} jointly trains a classifier and a generative adversarial network (GAN) \cite{goodfellow2014generative} where the classifier is trained to correctly classify the ID dataset but output a low confidence for the generated dataset. The generator is supervised not only by the discriminator but also by the classifier. This ensures that the generated dataset is neither too far nor too close to the training distribution. This is generally the essence of a synthetic OOD data; i.e., the generated data should approach the training distribution but not too close. \cite{vernekar2019outofdistribution} generates two types of OOD data using a conditional variational autoencoders (VAEs) \cite{kingma2022autoencoding}: samples that are close to the in-distribution but outside the in-distribution manifold and samples are in the in-distribution manifold but near the in-distribution boundary. The method trains a $K + 1$ classifier, where $K$ is the number of classes and the $K + 1-th$  class represents the OOD class. Virtual outlier synthesis (VOS) \cite{du2022vos} proposes to dynamically generate virtual outliers from  low-likelihood region of the Gaussian distribution formed from the empirical means and standard deviations of the features in the penultimate layer of the classification model. Another closely related work \cite{wang2023cmg} generates synthetic data by linearly interpolating the one-hot encodings of target classes in the training data to form pseudo class embeddings and generate an image by feeding the resulting embedding to the decoder network of a variational autoencoders (VAEs) \cite{kingma2022autoencoding}. This is similar to our data generation approach except we use diffusion models as opposed to VAEs.

\textit{Diffusion models for OOD detection}. Recently, diffusion models have been used for unsupervised anomaly detection \cite{mirzaei2022fake}, \cite{graham2023denoising}, \cite{liu2023unsupervised}. \cite{mirzaei2022fake} generates synthetic OOD data by early stopping, \cite{graham2023denoising} used the reconstruction error of a noised image as the score function, and \cite{liu2023unsupervised} intentionally corrupts the input image by cutting out a large portion of the image and reconstruct it using a pre-trained DDPM. The L2 distance between the original image and the reconstructed image is used as the OOD detection score function.





\section{The Proposed Method}
\label{chap:method}

Let $f_\theta(x)$ be a multi-class classification model where it takes an input $x \in \mathbb{R}^d$, and predicts a vector of probabilities $p_\theta(y|x)\in[0,1]^C$, where $d$ is the number of features and $C$ is the number of classes. Deep neural networks trained on a dataset $X=\{x_1,\dots,x_n\} \sim p_{data}(x)$ tend to make an overconfident prediction when exposed to previously unseen distribution, $p_{OOD}(x)$. Out-of-distribution detection aims to detect whether an input $x$ comes from $p_{data}(\cdot)$ or $p_{OOD}(\cdot)$.

Let $s_\theta(x) \in \mathbb{R}$ be a score function that assigns a higher value to in-distribution (ID) inputs and a lower value to out-of-distribution (OOD) inputs. The score function is used to measure how likely an input $x$ is to come from the training data distribution. If the score $s_\theta(x)$ is low, then the input $x$ is likely to be OOD data.

\subsection{Out-of-Distribution Data Generation using Diffusion Models}
To generate OOD data, we interpolate between the one-hot encoding vectors of any two different classes, which we refer to as label mixup. By doing so, we create new pseudo class embeddings that represent OOD data. These embeddings represent images that contain features from both classes and can be used to generate high-quality synthetic data. The resulting vector is mapped to the pixel space using a conditional DDPM pre-trained on the in-distribution data.

To explain why this works, we present an analogy between our proposed label mixup and mixup training \cite{zhang2018mixup}. Mixup training is a regularization (data augmentation) technique that trains a neural network on the convex combinations of pairs of examples and their labels. By doing so, mixup regularizes the neural network to favor simple linear behavior between training examples. The decision boundary in a model trained with mixup regularization smoothly decays from one class to another, thus predicting low confidence for data that lies in between. It has been shown that mixup training generalizes to out-of-distribution and adversarial examples. Similarly, label mixup can be considered as a form of mixup but in the label space instead of the image space.

To this end, we select any two different classes and add their one-hot encodings element-wise, and input the resulting label (in addition to a noise input sampled from a uniform distribution) to a pre-trained DDPM. To train the DDPM we use the pipieline from Hugging Face diffusers library\footnote{https://huggingface.co/docs/diffusers/index}. Interested readers can refer to \cite{ho2020denoising} for more details about diffusion models. See Figure \ref{fig:diffusion_label_mixup} for a snapshot of the generated data. The training scheme for the OOD detector will follow next.

\subsection{Deep Metric Learning}
The objective is to learn an encoder neural network $f_\theta$ such that samples of the same class are close in the embedding space while samples of different classes are far in the embedding space. In other words, we require small intra-class and a large inter-class variance. Suppose we have a neural network $f_\theta(x_i) \rightarrow z_i \in \mathbb{R}^d$, where $x_i$ is an input image and $z_i$ represents the features in the penultimate layer. Normally a fully connected layer with weights $W \in \mathbb{R}^{d\times C}$ and biases $b \in \mathbb{R}^C$ is used to project $z_i$ into the logit space and the softmax (cross-entropy) loss is derived as follows:

\begin{equation}
    \mathcal{L}_{\text{softmax}} = -\frac{1}{N} \sum^N_{i=1} \log \frac{e^{W^T_{y_i}z_i + b_{y_i}}}{\sum^C_{j=1} e^{W^T_{y_j}z_i + b_{y_j}}},
    \label{eqn:softmax}
\end{equation}
where $W_{y_i}, b_{y_i}$ are the weights and the bias associated with the class $y_i$, and $C$ is the number of classes. If we fix $b_{y_i} = 0$ and normalize the weights s.t. $|W_{y_i}| = 1$, we can rewrite $W^T_{y_j}z_i + b_{y_j}$ as $|z_i|\cos(\theta_{j,i})$, where $\theta_{j,i}$ is the angle between the feature vector $z_i$ and the class weights $W_{y_j}$. $|z_i|\cos(\theta_{j,i})$ is the projection of $z_i$ onto the class weights $W_{y_j}$. $x_i$ is classified as class $y_i$ if $\cos(\theta_{i,i}) > \cos(\theta_{j,i}), \forall j \in 1,\dots,C$. SphereFace \cite{liu2017sphereface} introduces a margin $m$ s.t. $x_i$ is classified as class $y_i$ if $\cos(m\theta_{i,i}) > \cos(\theta_{j,i}), \forall j \in 1,\dots,C$. This encourages a larger inter-class distance as it moves the decision boundary from a bisector between $W_i$ and $W_j$ to an angular margin $m$. The softmax loss with the proposed angular margin becomes:
\begin{equation}
    \mathcal{L}_{\text{angular}} = -\frac{1}{N} \sum^N_{i=1} \log \frac{e^{|z_i|\cos(m\theta_{i,i})}}{e^{ |z_i|\cos(m\theta_{i,i})} + \underset{j \ne i}{\sum} e^{ |z_i|\cos(\theta_{j,i})}},
    \label{eqn:sphereface}
\end{equation}
In this work, we regard state-of-the-art metric learning loss functions,  such as SphereFace \cite{liu2017sphereface}, CosFace \cite{wang2018cosface}, ArcFace \cite{deng2020sub}, and AdaCos \cite{zhang2019adacos} as OOD score functions. We will briefly describe each loss function. A comprehensive analysis of each loss function is beyond the scope of this paper, and readers are encouraged to refer to the respective papers for detailed derivations. For our purposes, we treat the loss functions the same; we use the maximum cosine similarity between the features $z_i$ and the weight vectors $W_i, \forall i \in 1,\dots,C$, as the OOD score function. In addition to normalizing the weight vectors $W_{y_i}, \forall i=1,\dots,C$, CosFace \cite{wang2018cosface} normalizes the features vectors $z_i$ s.t. $|z_i| = 1$. The CosFace loss function is defined as follows:
\begin{equation}
    \mathcal{L}_{\text{CosFace}} = -\frac{1}{N} \sum^N_{i=1} \log \frac{e^{s(\cos(\theta_{i,i}) -m )}}{e^{ s(\cos(\theta_{i,i}) - m)} + \underset{j \ne i}{\sum} e^{ s\cos(\theta_{j,i})}},
    \label{eqn:cosface}
\end{equation}
where $s$ is a scaling factor and $m$ is the margin. ArcFace \cite{deng2020sub}, like CosFace, normalizes both the weights and the features. However, the angular margin is defined in the angle space as given in the following equation:
\begin{equation}
    \mathcal{L}_{\text{ArcFace}} = -\frac{1}{N} \sum^N_{i=1} \log \frac{e^{s\cos(\theta_{i,i} + m)}}{e^{ s\cos(\theta_{i,i} + m)} + \underset{j \ne i}{\sum} e^{ s\cos(\theta_{j,i})}},
    \label{eqn:arcface}
\end{equation}
Adacos \cite{zhang2019adacos} proposed a fixed scaling parameter $s$ defined by the following equation:
\begin{equation}
    s \approx \sqrt{2}\log(C - 1),
\end{equation}
where $C$ is the number of classes. The loss function remains the sames as Eqn. (\ref{eqn:arcface}).

\subsection{OOD Detector Training}
In addition to the synthetic data generation approach, our contribution is using the deep metric learning loss functions for training OOD detectors. We train two types of models: with and without synthetic outlier exposure. Both types of models are trained with the vanilla softmax loss function (Eqn. (\ref{eqn:softmax})) and the metric learning-based loss functions (e.g., Eqn. (\ref{eqn:sphereface})).

Metric learning-based OOD detection have been studied before in \cite{ravikumar2023intra}, and \cite{techapanurak2020hyperparameter}. \cite{techapanurak2020hyperparameter} uses the scaled cosine similarity as the score function. Specifically, the weights and features are normalized to be a unit vector and their dot product scaled by a learnable parameter $s$ is used as the score function, i.e., $\cos(\theta_{i,j}) = W^T_{y_j} z_i/|W_{y_j}||z_i|$, and $\mathcal{L} = -\frac{1}{N}\sum^N_i \log\frac{e^{s\cos(\theta_{i,i})}}{\sum^C_{j=1} s\cos(\theta_{i,j})}$, where $z_i$ represents the features of input $x_i$ and $W_{y_j}$ are the weights of the class $j-th$ class. In this study, we aim to explore more metric learning loss functions including the scaled cosine loss function and compare their OOD detection performance before and after the proposed outlier exposure.

\subsection{Detecting OOD samples}
A test sample $x$ is predicted as OOD if the maximum cosine similarity between the normalized features and weights is less than a threshold $\tau$, otherwise it is predicted as ID. The threshold is computed from a validation set at 95\% true positive rate (TPR). If $x$ is predicted as ID, we predict its class as the index with the maximum cosine similarity scaled by $s$ and $m$. For example, when using the SphereFace loss function (Eqn. (\ref{eqn:sphereface})), the index with the highest $\cos(m\theta_{i,i})$ becomes the predicted class. For models trained with the softmax loss function, the maximum probability used to decide if $x$ is ID or OOD.

\section{Experiments}
We use the ResNet-50 \cite{he2016deep} architecture to train all models. We use CIFAR-10 datasets as the in-distribution dataset for all experiments and we use CIFAR-100, Tiny ImageNet \cite{le2015tiny}, SVHN \cite{netzer2011reading}, iSUN \cite{xu2015turkergaze}, LSUN \cite{yu2015lsun} as out-of-distribution datasets.

\subsection{Experimental Results}
We compare the AUROC, AUPR-In (when the in-distribution data is the positive class) and AUPR-Out (when the OOD data is the positive class) results of the baseline models and our models in Table \ref{tab:main_results}. We use the model trained with the softmax loss function and the MSP \cite{hendrycks2016baseline} score function as a baseline and compare it against our models trained with the metric learning-based loss functions. In Table \ref{tab:main_results} (top), we show the models before using synthetic outlier exposure and the performance after synthetic outlier exposure in Table \ref{tab:main_results} (bottom). The models with the synthetic data show a significant performance gain in both the vanilla softmax and the metric learning loss functions. Notably, when the baseline models including the softmax loss function and the metric loss functions struggle with certain datasets such as Gaussian noise, uniform noise, and Tin (R), the models with the outlier exposure produce consistent results across all datasets. Furthermore, the scaled cosine (which is also metric learning-based loss function) outperforms the softmax based training.

In summary, the results show that the proposed data generation approach generalizes across several training loss functions, usually with a significant improvement.

\begin{table*}
  \centering
  \begin{adjustbox}{width=\textwidth}
    \begin{tabular}{c|l|c|c|c|c|c|c}
      \toprule
      \toprule
      \multirow{2}{*}{\rotatebox[origin=c]{90}{\textbf{Without Outlier Exposure}}} & Method & {Softmax} & {Scaled Cosine} & {AdaCos (ours)} & {ArcFace (ours)} & {CosFace (ours)} & {SphereFace (ours)} \\
      \cmidrule{2-8}
      & CIFAR-100 & \textbf{0.913}/\textbf{0.896}/\textbf{0.892} & 0.869/\underline{0.871}/0.868 & \underline{0.885}/0.868/\underline{0.884} & 0.878/0.858/0.878 & 0.844/0.806/0.856 & 0.874/0.851/0.879 \\
      & LSUN (C) & 0.937/0.917/0.928 & \textbf{0.977}/\textbf{0.977}/\textbf{0.978} & 0.969/0.965/0.970 & 0.966/0.963/0.967 & 0.965/0.960/0.968 & \underline{ 0.974}/\underline{0.973}/\underline{0.975} \\
      & LSUN (R) & \textbf{0.962}/\textbf{0.967}/\underline{0.949} & \underline{0.961}/\underline{0.963}/\textbf{0.961} & 0.942/0.939/0.942 & 0.937/0.931/0.939 & 0.943/0.945/0.942 & 0.941/0.935/0.946 \\
      & Tin (C) & \underline{0.960}/\underline{0.964}/\underline{0.945} & \textbf{0.965}/\textbf{0.968}/\textbf{0.963} & 0.928/0.928/0.927 & 0.912/0.898/0.920 & 0.931/0.921/0.934 & 0.934/0.929/0.938 \\
      & Tin (R) & \textbf{0.950}/\textbf{0.952}/\underline{0.932} & \underline{0.945}/\underline{0.950}/0.942 & 0.880/0.872/0.883 & 0.858/0.833/0.873 & 0.895/0.885/0.896 & 0.886/0.874/0.896 \\
      & iSun & \textbf{0.963}/\textbf{0.971}/\underline{0.944} & \underline{0.956}/\underline{0.961}/\textbf{0.952} & 0.925/0.925/0.920 & 0.919/0.915/0.917 & 0.940/0.946/0.932 & 0.934/0.935/0.932 \\
      & SVHN & 0.955/0.925/0.973 & 0.966/0.930/\underline{0.986} & \textbf{0.978}/\textbf{0.951}/\textbf{0.991} & \underline{0.968}/0.943/\underline{0.986} & 0.951/0.853/0.982 & \underline{0.968}/\underline{0.947}/0.985 \\
      & Gaussian Noise & 0.837/0.878/0.700 & \textbf{1.000}/\textbf{1.000}/\textbf{1.000} & \underline{0.999}/\underline{0.999}/\underline{0.998} & 0.999/0.999/0.997 & \textbf{1.000}/\textbf{1.000}/\textbf{1.000} & 0.999/0.999/0.997 \\
      & Uniform Noise & 0.923/0.951/0.847 & \textbf{1.000}/\textbf{1.000}/\textbf{1.000} & 1.000/1.000/1.000 & 0.999/0.999/0.996 & \textbf{1.000}/\textbf{1.000}/\textbf{1.000} & \textbf{1.000}/\textbf{1.000}/\underline{0.998} \\
      \toprule
      \toprule
      \multirow{2}{*}{\rotatebox[origin=c]{90}{\textbf{With Outlier Exposure}}} & CIFAR-100 & \underline{0.919}/\underline{0.919}/\underline{0.902} & \textbf{0.934/0.935/0.929} & 0.894/0.885/0.883 & 0.902/0.895/0.891 & 0.902/0.906/0.891 & 0.888/0.885/0.879 \\
      & LSUN (C) & 0.975/\textbf{0.980}/0.970 & \textbf{0.983/0.985/0.982} & 0.972/0.975/0.971 & \textbf{0.976}/0.978/\textbf{0.974} & 0.974/0.976/0.973 & 0.969/0.972/0.968 \\
      & LSUN (R) & \underline{0.984}/\underline{0.986}/\underline{0.983} & \textbf{0.990/0.991/0.990} & 0.978/0.981/0.975 & 0.981/0.984/0.980 & 0.972/0.975/0.970 & 0.973/0.975/0.972 \\
      & Tin (C) & \underline{0.974}/\underline{0.979}/\underline{0.968} & \textbf{0.985/0.987/0.983} & 0.955/0.960/0.948 & 0.967/0.973/0.962 & 0.961/0.966/0.957 & 0.944/0.945/0.942 \\
      & Tin (R) & \underline{0.973}/\underline{0.977}/\underline{0.968} & \textbf{0.984/0.985/0.983} & 0.953/0.956/0.948 & 0.965/0.969/0.960 & 0.952/0.958/0.947 & 0.935/0.933/0.935 \\
      & iSun & \underline{0.980}/\underline{0.985}/\underline{0.975} & \textbf{0.988/0.990/0.987} & 0.968/0.976/0.960 & 0.974/0.980/0.967 & 0.965/0.972/0.958 & 0.962/0.968/0.957 \\
      & SVHN & 0.962/0.943/0.978 & \textbf{0.985/0.974/0.993} & 0.964/0.936/0.983 & \underline{0.979}/\underline{0.957}/\underline{0.957} & 0.969/0.944/0.986 & 0.976/0.954/0.989 \\
      & Gaussian Noise & 0.976/0.985/0.951 & \textbf{0.999/0.999/0.999} & \textbf{0.999/0.999/0.999} & \textbf{0.999/0.999}/0.996 & 0.999/0.999/0.995 & \textbf{0.999/0.999}/\underline{0.997} \\
      & Uniform Noise & 0.985/0.991/0.968 & \textbf{0.999/1.000/0.999} & \textbf{0.999}/0.999/\textbf{0.999} & \textbf{0.999/0.999}/0.996 & 0.999/0.999/0.999 & \textbf{0.999/0.999}/\underline{0.997} \\
      \toprule
      \toprule
      \end{tabular}
  \end{adjustbox}
  \caption{Out-of-distribution detection evaluation results before and after outlier exposure. The numbers separated by / indicate AUROC/AUPR-In/AUPR-Out. \textbf{Boldface} indicates the best approach and underline (\_) indicates the second best. The amount of increase/decrease from the baseline (without outlier exposure) to the models trained with outlier exposure is indicated in the last section (difference) of this table.}
  \label{tab:main_results}
\end{table*}

\paragraph{Closed-Set Accuracy} We compare the in-distribution classification accuracy of the baseline models (i.e., trained on ID data only) and their performance after synthetic outlier exposure. Table \ref{tab:closed_set_accuracy} illustrates that the OOD detectors' closed-set accuracy is comparable to that of the baseline classifiers. Our models have the ability to detect out-of-distribution samples with a small drop in closed-set classification accuracy.

\begin{table*}
  \centering
  \begin{adjustbox}{width=\textwidth}
    \begin{tabular}{l|c|c|c|c|c|c}
      \hline
      & Softmax & Scaled Cosine \cite{techapanurak2020hyperparameter} & AdaCos \cite{zhang2019adacos} & ArcFace \cite{deng2020sub} & CosFace \cite{wang2018cosface}  & SphereFace \cite{liu2017sphereface}\\
      \hline
      Standard & \textbf{96.98} & 92.28 & \textbf{96.39} & \textbf{96.41} & \textbf{96.08} & \textbf{96.21}\\
      \hline
      Outlier Exposure (ours) & 96.54 & \textbf{96.78} & 95.80 & 95.81 & 95.61 & 95.70\\
      \hline
    \end{tabular}
  \end{adjustbox}
  \caption{Our proposed outlier exposure has minimal impact on the in-distribution classification accuracy as it remains largely unchanged. We highlight in \textbf{bold} the model with the higher in-distribution accuracy for each loss function.}
  \label{tab:closed_set_accuracy}
\end{table*}

\section{Conclusion}
In conclusion, this paper introduced a novel method for out-of-distribution (OOD) detection in classification models by combining deep metric learning and synthetic data generation using diffusion models. The approach employs outlier exposure, a popular technique for OOD detection, where models are trained using known OOD samples. During training, the model low confidence for the training OOD data, while accurately classifying the in-distribution (ID) data. To generate synthetic OOD data, we proposed a label-mixup approach using Denoising Diffusion Probabilistic Models (DDPMs), and we utilize recent advancements in metric learning to train our models.

The experimental results demonstrate that our method, employing outlier exposure with metric learning, outperforms softmax training in most settings. Moreover, all loss functions including the vanilla softmax and metric learning-based loss functions show a significant improvement after the proposed outlier exposure.

The experimental results demonstrate that our method, employing outlier exposure shows a significant improvement across a range of loss functions and datasets compared to the baseline models that do not have access to a training OOD data.

{\small
\bibliographystyle{ieee_fullname}
\bibliography{reference}
}

\end{document}